\documentclass[conference]{IEEEtran}
\usepackage{cite}
\usepackage{amsmath,amssymb,amsfonts}
\usepackage{algorithmic}
\usepackage{graphicx}
\usepackage{textcomp}
\usepackage{xcolor}
\usepackage{tabularx}

\newcommand{\ie}{i.e.,}

\usepackage[breaklinks,colorlinks]{hyperref}
\usepackage{booktabs}
\newcolumntype{C}{>{\centering\arraybackslash}X}

\newcolumntype{H}{>{\setbox0=\hbox\bgroup}c<{\egroup}@{}}
\def\BibTeX{{\rm B\kern-.05em{\sc i\kern-.025em b}\kern-.08em
    T\kern-.1667em\lower.7ex\hbox{E}\kern-.125emX}}
\begin{document}

\title{Pyramid-based Mamba Multi-class Unsupervised Anomaly Detection}

\author{\IEEEauthorblockN{Nasar Iqbal\IEEEauthorrefmark{1},
Niki Martinel\IEEEauthorrefmark{2}}
\IEEEauthorblockA{Department of Mathematics, Computer Science and Physics, University of Udine,
Udine, Italy\\
Email: \IEEEauthorrefmark{1}iqbal.nasar@spes.uniud.it,
\IEEEauthorrefmark{2}niki.martinel@uniud.it}
}

\maketitle

\begin{abstract}
Recent advances in convolutional neural networks (CNNs) and transformer-based methods have improved anomaly detection and localization, but challenges persist in precisely localizing small anomalies.
While CNNs face limitations in capturing long-range dependencies, transformer architectures often suffer from substantial computational overheads.
We introduce a state space model (SSM)-based Pyramidal Scanning Strategy (PSS) for multi-class anomaly detection and localization--a novel approach designed to address the challenge of small anomaly localization.
Our method captures fine-grained details at multiple scales by integrating the PSS with a pre-trained encoder for multi-scale feature extraction and a feature-level synthetic anomaly generator.
An improvement of $+1\%$ AP for multi-class anomaly localization and a +$1\%$ increase in AU-PRO on MVTec benchmark demonstrate our method's superiority in precise anomaly localization across diverse industrial scenarios. The code is available at \href{https://github.com/iqbalmlpuniud/Pyramid_Mamba}{https://github.com/iqbalmlpuniud/Pyramid\_Mamba}.

\end{abstract}


\section{Introduction}
\label{sec:intro}

Visual anomaly detection (AD) is a critical task in computer vision with direct implications for safety, quality control, and security.
Anomalies, defined as deviations from expected patterns, often signal defects in manufacturing, security breaches, or early signs of disease in medical images. This task is challenged by the need to accurately capture a wide range of anomaly types and scales and the computational demands of processing images in real-time.
Overcoming these challenges is essential for developing reliable automated systems that reduce human intervention and errors.
Advances in AD can significantly enhance operational efficiency, improve product quality, and ensure safer outcomes across various applications.

Current techniques addressing the AD challenges can be categorized into three types: embedding, synthesizing, and reconstruction-based approaches. Embedding methods leverage ImageNet pre-trained models for extracting features to obtain an anomaly-free feature distribution through statistical methods
\cite{Defard2021,Roth2022}.
These methods often struggle in many settings due to the ImageNet data domain mismatch.
Synthesizing approaches (\textit{e.g.},~\cite{Li2021}) generate artificial anomalies for training, yet face challenges in producing realistic defects.
Reconstruction-based techniques (\textit{e.g},~\cite{He2024,Deng2022}), typically using autoencoders, may inadvertently learn to reconstruct anomalies, reducing their efficacy.
Among these methods, Reconstruction-based approach demonstrates superior performance compared to Synthesizing and Embedding-based approaches for AD. However, while it leverages CNN to effectively capture local context, it struggles to capture long-range dependencies. Transformer-based approaches were introduced to overcome such limitations.
Still, their quadratic computational complexity \cite{You2022} results in scalability issues, restricting their use to smallest
feature maps for anomaly detection.
This underscores the
need for more efficient architectures that can effectively
capture long-range feature dependencies while maintaining
computational feasibility. 
Recently, several studies \cite{Liu2024, Shi2024, Ruan2024, Wang2024, Huang2024} have successfully integrated state-space models (SSMs), particularly Mamba \cite{Gu2023}, into computer vision tasks due to their considerably low linear complexity as compared to transformers, while delivering improved performance.
MambaAD \cite{Huang2024} exploited mamba for AD.
While computationally efficient, its reliance on a single-scale SSM limits its ability to localize anomalies at different granularities. This limitation underscores the need of developing a multi-scale approach capable of operating at multiple levels of granularity to effectively detect small, localized anomalies.
Building on this insight, we introduce a novel Pyramidal Scanning Strategy (PSS) that leverages the image pyramid to enhance global and local dependencies, improving anomaly localization accuracy.

While Mamba excels at modeling global dependencies, it lacks the ability to capture local spatial features, which are crucial for detecting pixel-level anomalies. To bridge this gap, we introduce a Context-aware State Space (CSS) module, that integrates mamba for modeling global dependencies and convolutional layers for modeling local dependencies. 
This hybrid approach ensures robust feature representation by integrating local context with global dependencies, producing precise anomaly maps across multiple scales. In addition, small anomalies are subtle and often localized within specific regions, making them difficult to detect at a single resolution. To address this, our novel PSS leverages a pyramid-based processing scheme, that progressively downscales the image, creating multiple scales of representation.
This allows the model to capture small, localized defects and broader anomalies, enhancing the overall detection accuracy.
 

The key contributions of this work are as follows:
\begin{itemize}
    \item We propose a Context-aware State Space (CSS) module that integrates convolutional layers to capture fine-grained local dependencies, complementing the global modeling capacity of the mamba, thus ensuring robust feature representation across scales.   
    
    \item We introduce a novel Pyramidal Scanning Strategy (PSS) that leverages multi-scale image pyramids to enhance the detection of both small, localized anomalies and large, global structural deviations while maintaining low computational complexity.
    
    \item Comprehensive evaluations across multiple unsupervised AD tasks demonstrate that our pyramid-based Mamba model achieves state-of-the-art anomaly detection and localization performance at image and pixel levels.
\end{itemize}


\begin{figure*}[t]
  \centering
  \includegraphics[width=0.80\linewidth]{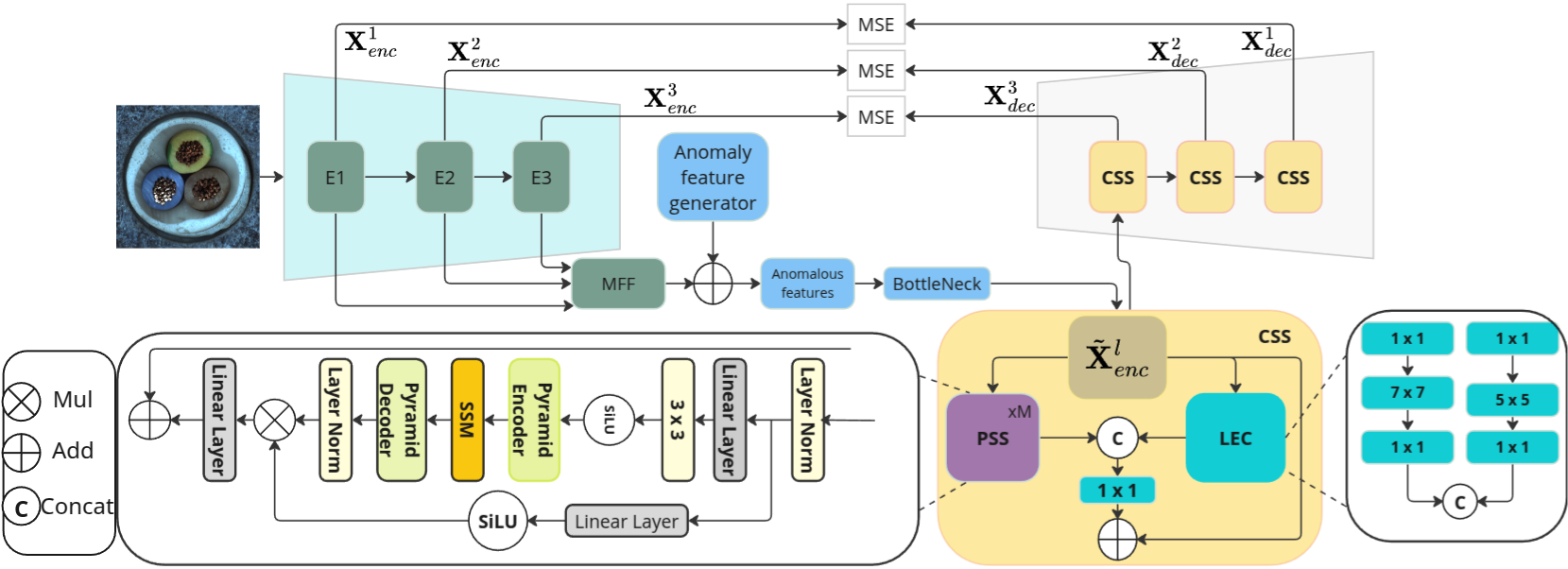}
  \caption{Overview of the proposed Pyramid-Mamba approach consisting of a teacher-student network to reconstruct multi-scale synthetic anomalous features. Each Context-aware State Space (CSS) module consists of Pyramidal Scanning Strategy (PSS) to capture local and global interaction and parallel multi-kernel convolution operations to capture local information. An anomaly map is the sum of multi-scale reconstruction errors.}
  \label{fig:mamba}
  \vspace{-1em}
\end{figure*}


\section{Related Work}
\subsection{Unsupervised Anomaly Detection}
Several approaches have been developed to address anomaly detection and localization in industrial applications.
These methods can be broadly divided into Single-class and Multi-class-Anomaly detection. 

\subsubsection{Single-class Anomaly detection}

\textbf{Reconstruction-based methods} propose models that are trained to reconstruct anomaly-free images, with anomalous images used only during inference. 
The underlying principle is that regions containing anomalies will be poorly reconstructed due to their absence during training.
Techniques such as auto-encoders \cite{Gong2019} and generative adversarial models \cite{Gu2023} were employed to reconstruct anomaly-free images during training.
Anomaly maps are generated by comparing input and reconstructed images, providing scores for detection and localization.
However, these methods may struggle when anomalous and non-anomalous regions share common patterns \cite{Zavrtanik2021}.

\textbf{Synthesizing-based methods} integrate artificial anomalies within anomaly-free images. In~\cite{Zavrtanik2021}, pseudo anomalies are generated using Perlin noise from texture images during training to obtain a model predicting the anomaly masks. 


\textbf{Embedding-based methods} extract low-dimensional representations of anomaly-free images using ImageNet pre-trained networks, then detect anomalies by measuring deviations from this learned distribution in the embedded space. Patch-level features embedding via multivariate Gaussian distributions \cite{Defard2021}, and memory banks with Mahalanobis distance for comparison \cite{Roth2022} were explored to enhance discrimination between normal and anomalous samples.
Knowledge distillation techniques have also gained traction, where student networks are trained to mimic the features of fixed, pre-trained teacher networks \cite{Deng2022}, facilitating efficient feature learning for anomaly detection.


\subsubsection{Multi-class Anomaly Detection}
Multi-class AD methods address the fundamental challenge of efficiently handling diverse classes with a single model while mitigating excessive computational and memory demands.
The emergence of Multi-class Unsupervised Anomaly Detection (MUAD) techniques \cite{Huang2024} has marked a significant step towards properly handling these issues. Recent contributions include a comprehensive multi-class framework \cite{You2022} and a vector quantization approach that mitigates the learning of "identical shortcuts" \cite{Lu2023}. 

Presently, the majority of the AD approaches utilize a single-class framework. In this approach, individual model is trained and tested for each class which leads to significant increase in training and memory usage. Although multiclass settings have proven effective in addressing this issue, these methods can suffer from different challenges, such as \cite{Huang2024}, \cite{Lu2023} utilize transformers for global modeling, and their performance is impacted by the quadratic computational complexity. On the other hand, MambaAD \cite{Huang2024} successfully, leverages the lower linear complexity of mamba compared to transformers. However, it uses single SSM for global modeling which limits its ability to detect small, localize anomalies.
Our approach extends these efforts by introducing a pyramidal scanning strategy that effectively captures multiscale information, enabling more accurate anomaly detection and localization across diverse classes while maintaining computational efficiency.

\subsection{State Space Models}
State Space Models (SSMs) \cite{Gu2023,Gu2021,Smith2022,Mehta2022} have proven effective for long-sequence modeling, providing efficient handling of long-range dependencies. The seminal Structured State-Space Sequence (S4) model \cite{Gu2021} introduced a novel diagonal structure for parameterization, enabling linear complexity in sequence modeling. Building on this, models like S5 \cite{Smith2022}, H3 \cite{Gong2019}, and Mamba \cite{Gu2023} have further improved SSMs' capabilities. 
Mamba proposed a data-driven selection method within S4 that maintains linear complexity for long-sequence modeling, which has inspired several computer vision applications \cite{Ruan2024, Wang2024}.
 Recently, Mamba was also adapted for AD~\cite{Huang2024}, which uses a single SSM to model the entire image, limiting its ability to localize anomalies at different granularities.

In contrast, our method introduces a pyramid-based scanning mechanism, significantly enhancing anomaly localization by operating at multiple levels granularity .
Each pyramid level provides a finer resolution, allowing for better detection of small, localized anomalies and captures global structural patterns.
As shown by experimental results, our novel multi-scale approach enables more precise anomaly localization than~\cite{Huang2024}.


\section{Method}
The pipeline of our method is shown in~\figurename~\ref{fig:mamba}.
It comprises three main components: a pre-trained encoder block, a bottleneck layer for feature fusion, and a pyramid-based Mamba decoder.
The input image $\mathbf{I} \in \mathbb{R}^{H \times W \times 3}$ is processed by the encoder to extract features at multiple levels, denoted as $\mathbf{X}_{\text{enc}}^{l} \in \mathbf{R}^{H^l \times W^l \times C^l}$ for each level $l$.
These multi-level features are downsampled through one or more $3 \times 3$ convolutions with a stride of 2 to match sizes before being fed to the Multi-Scale Feature Fusion (MFF) block combining low and high resolution feature maps.
The resulting features are then perturbed with synthetic noise to get  $\mathbf{\tilde{X}}_{\text{enc}}^{l}$. 
At each level of the pyramid-based decoder, we introduce a Context-aware State Space (CSS) module that incorporates our Pyramidal Scanning Strategy  (\texttt{PSS}).
This enables multi-directional scanning across different pyramid levels, ensuring that global and local information is captured effectively. 
The decoder is trained to minimize the reconstruction error between the encoder features $\mathbf{X}_{\text{enc}}^{l}$ and the corresponding decoder reconstructions $\mathbf{X}_{\text{dec}}^{l} \in \mathbb{R}^{H^l \times W^l \times C^l}$, with inter-level communication facilitating more accurate feature recovery.
To obtain a 2D anomaly map we adopted the same approach in~\cite{Deng2022} with multi-scale averaging.

\subsection{Background}
SSMs can be considered linear time-invariant systems that represent a class of sequence models that maps a one-dimensional input stimulation \( x(t) \in \mathbb{R} \) to response \( y(t) \in \mathbb{R} \) via hidden space \( h(t) \in \mathbb{R}^N \). SSMs can be expressed using linear ordinary differential equations as: 
\begin{equation}
h'(t) = \mathbf{A}h(t) + \mathbf{B}x(t), \quad y(t) = \mathbf{C}h(t),
\end{equation}
where \( \mathbf{A} \in \mathbb{R}^{N \times N} \), \( \mathbf{B} \in \mathbb{R}^{N \times 1} \), and \( \mathbf{C} \in \mathbb{R}^{1 \times N} \) are state transition matrix. 

To be practically implemented, the continuous-time SSMs must be discretized in advance. (1) is commonly  discretized using zero-order hold (ZOH) which as expressed as:
\begin{equation}
\mathbf{\bar{A}} = \exp(\Delta \mathbf{A}), \quad \mathbf{\bar{B}} = (\Delta \mathbf{A})^{-1} (\exp(\Delta \mathbf{A}) - I) \cdot \Delta \mathbf{B}.
\end{equation}
After discretization, SSM-based models can be formulated as: 
\begin{equation}
h_t = \mathbf{\bar{A}} h_{t-1} + \mathbf{\bar{B}} x_t, \quad y_t = \mathbf{C} h_t.
\end{equation}

Such a formulation can be translated into a global convolutional operator computing 
\begin{equation}
\mathbf{K} = (\mathbf{C}\mathbf{\bar{B}}, \mathbf{C}\mathbf{\bar{A}}\mathbf{\bar{B}}, \ldots, \mathbf{C}\mathbf{\bar{A}}^{L-1}\mathbf{\bar{B}}), \quad y = \mathbf{x} \ast \mathbf{K}.
\end{equation}
where $\ast$ represents convolution operation, \( \mathbf{K} \in \mathbb{R}^L \) is an SSM kernel, and \( L \) is the sequence length of input $\mathbf{x} = [x(0), \cdots, x(L)]$.

\subsection{Context-aware State Space (CSS)}
Effective AD requires capturing both long-range dependencies and fine-grained local details.
CNNs, while efficient in modeling local features, struggle to capture long-range dependencies. Transformers, though proficient in global modeling via self-attention, suffer from high computational complexity. To address these challenges, we introduce the CSS module leveraging the strengths of global and local modeling via the novel Pyramidal Scanning Strategy    (\texttt{PSS}) and the Locality-Enhanced Convolutional (\texttt{LEC}) blocks.

\textbf{The PSS} captures global context through multi-scale scanning.
For an input tensor $\mathbf{\tilde{X}}_{\text{enc}}^l$, the \texttt{PSS} operates at multiple pyramid levels, producing global features
\begin{equation}
\mathbf{X}_{\text{g}} = \texttt{PSS}_M(\cdots\texttt{PSS}_1(\texttt{PSS}_0(\mathbf{\tilde{X}}_{enc}^l))\cdots),
\end{equation}
with $M$ denoting the number of sequential PSSs.

\textbf{Locality-Enhanced Convolutional Block (LEC)} complements the global modeling by capturing local information using depth-wise convolutions.
Local features are obtained via two parallel blocks, each computing 
\begin{equation}
    \mathbf{X}_{\text{c},k} = \texttt{Conv}_{1 \times 1}(\texttt{DWConv}_{k \times k}(\texttt{Conv}_{1 \times 1}(\mathbf{\tilde{X}}_{enc}^l))),
\end{equation}
where $k\in\{5, 7\}$ represents the kernel size, allowing for fine-grained feature extraction with low computational cost.

\textbf{PSS and LEC feature fusion} is obtained via 
\begin{equation}
    \mathbf{X}^{\text{css}} = \texttt{Conv}_{1 \times 1}(\texttt{Concat}(\mathbf{X}_{\text{g}}, \mathbf{X}_{\text{c},5}, \mathbf{X}_{\text{c},7})) + \mathbf{\tilde{X}}_{enc}^l,
\end{equation}
where the residual connection is included to improve gradient flow, enhancing stability and convergence.

\subsection{Pyramidal Scanning Strategy (PSS)}

\begin{figure}[t]
  \centering
  
   \includegraphics[width=0.30\textwidth]{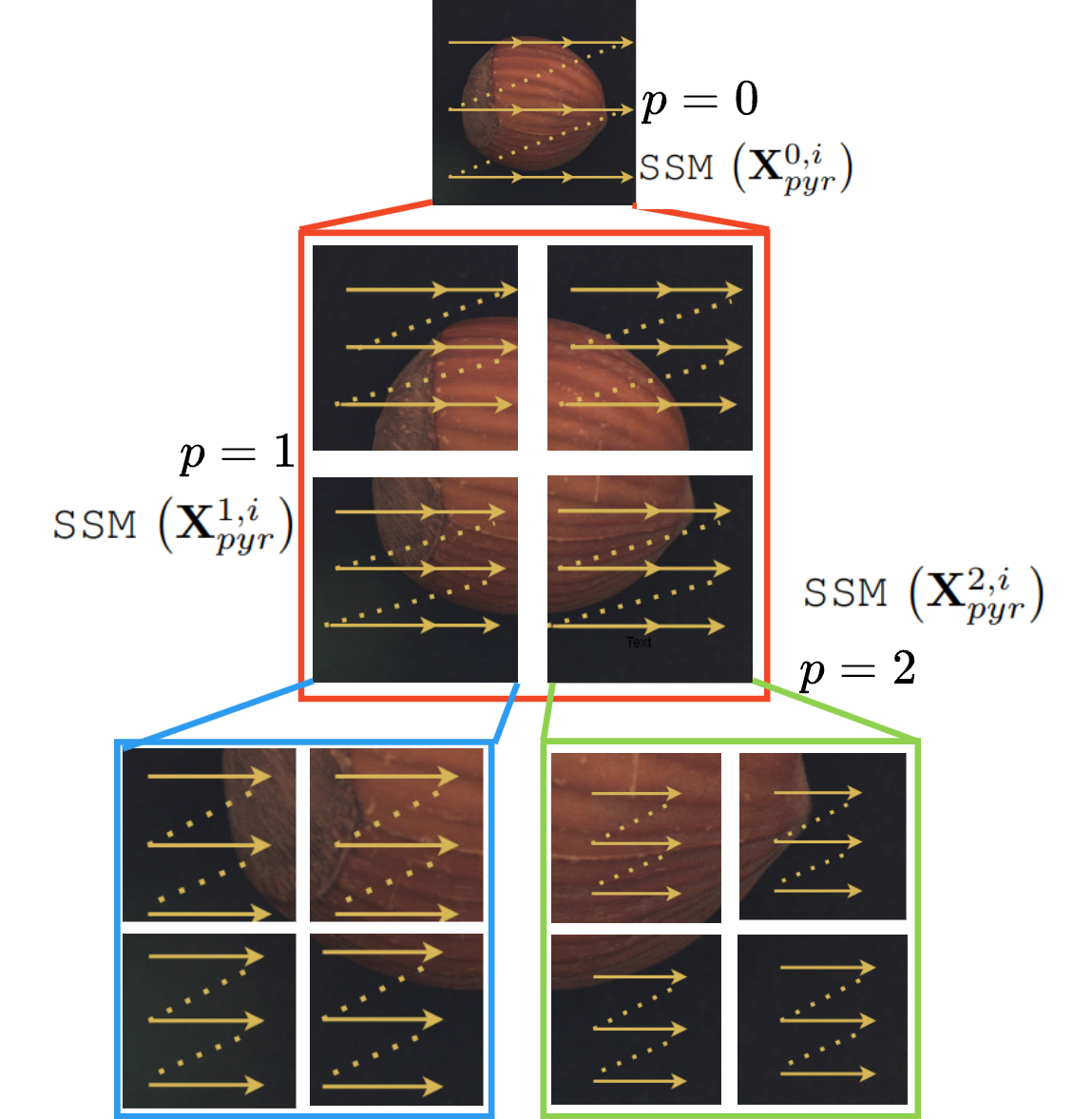}

   \caption{The proposed pyramid-based scanning involves applying a Selective Scan Module (SSM) to the entire image, then dividing it into four primary patches, and each processed separately by SSM. Each primary patch is further subdivided into four sub-patches and each is processed by SSM separately.}
   

   \label{fig:pyramid}
\end{figure}

The PSS is introduced to overcome the limitations of standard sequential and convolutional models in effectively capturing both local and global spatial dependencies in feature maps.
The PSS constructs a spatial pyramid representation of $\mathbf{\tilde{X}}_{enc}^l$ by recursively applying an SSM across multiple scales, 
enabling coarse and fine-grained feature extraction.

Let $\mathbf{X}^{p}_{pyr} \in \mathbb{R}^{H^{p,i} \times W^{p,i} \times C^{p,i}}$ be the input at pyramid level $p \in \{0, \dots, P\}$, where $P$ is the total number of pyramid levels.
For such an input, the $p$-th pyramid level computes
\begin{align}   
\mathbf{\hat{X}}^{p}_{pyr} &= \texttt{PS}(\mathbf{X}^{p}_{pyr}) \\
&= \texttt{cat} \left( \left\{ \texttt{SSM} \left( \mathbf{X}_{pyr}^{p,i} \right) \middle| \mathbf{X}_{pyr}^{p,i} \in \texttt{split} \left( \mathbf{X}^{p}_{pyr}, p \right) \right\} \right)\nonumber,
\end{align}
where $
\texttt{split}
\left( \mathbf{X}_{pyr}^{p}, p \right)
\to \left\{ \mathbf{X}_{pyr}^{p,i}, \right\}_{i=1}^{2^{p} \times 2^{p}}
$ yields $\mathbf{X}_{pyr}^{p,i}$ representing the $i$-th input sub-region of size $H/2^p \times W/2^p \times C$ at level $p$.

To capture local dependencies within the input, each sub-region $\mathbf{x}_i$ is independently analyzed through an $\texttt{SSM}(\cdot)$ layer applying the four-way SSM strategy in \cite{Liu2024}.
$\texttt{cat}(\cdot)$ organizes the feature maps back into their original grid structure, maintaining the spatial dimensions.
Due to such a reorganization, the shape of $\mathbf{\hat{X}}_{pyr}^{p}$ is the same as $\mathbf{X}_{pyr}^{p}$.

As shown in~\figurename~\ref{fig:pyramid}, we recursively apply the $\texttt{split}(\cdot)$ and $\texttt{SSM}(\cdot)$ operators to every $\mathbf{X}_{pyr}^{p,i}$ to ensure that both local details and broader contextual information are captured across the pyramid levels.
At the output of each recursive process, the $i$-th sub-region SSM output at level $p$, \ie $\texttt{SSM}(\mathbf{X}_{pyr}^{p,i})$, is averaged with the output of the $\texttt{PS}(\mathbf{X}^{p,i}_{pyr})$.



\begin{table*}[htbp]
\caption{Comparison with SoTA methods on \textbf{MVTec-AD} dataset for multi-class anomaly detection with AU-ROC/AP/$F1_{\text{max}}$}
\begin{center}
\begin{tabular}{c|ccccc}
\toprule
\textbf{Class} & \textbf{DeSTeg~\cite{Zhang2023}} & \textbf{RD4AD~\cite{Deng2022}} & \textbf{DiAD~\cite{He2024}} & \textbf{MambaAD~\cite{He2024arXiv}}  & \textbf{Pyramid-Mamba~(Ours)} \\
\midrule
Bottle & 98.7/99.6/96.8 & 99.6/99.9/98.4 & 99.7/96.5/91.8 & \textbf{100.0/100.0/100.0}  & \textbf{100.0/100.0/100.0} \\
Cable & 89.5/94.6/85.9 & 84.1/89.5/82.5 & 94.8/98.8/95.2 & 98.8/99.2/95.7  & \textbf{99.2/99.5/96.8} \\
Capsule & 82.8/95.9/92.6 & 89.0/97.5/95.5 & 94.4/98.7/94.9 &94.1/96.9/\textbf{96.9}  & \textbf{95.1/99.0}/94.8 \\
Hazelnut & 98.8/99.2/98.6 & 60.8/69.8/86.4 & 99.5/99.7/97.3 & \textbf{100.0/100.0/100.0}  & \textbf{100.0/100.0/100.0} \\
Metal Nut & 92.9/98.4/92.2  & \textbf{100.0/100.0/99.5} & 99.1/96.0/91.6 & 99.9/100.0/99.5 & 99.7/99.9/99.5 \\
Pill & 77.1/94.4/91.7 & \textbf{97.5}/\textbf{99.6}/\textbf{96.8} & 95.7/98.5/94.5 & 97.0/99.5/96.2  & 96.9/99.5/96.6 \\
Screw & 69.9/88.4/85.4 & \textbf{97.7}/99.3/95.8 & 90.7/\textbf{99.7}/97.9 & 94.7/97.9/94.0  & 96.9/99.5/\textbf{96.6} \\
Toothbrush & 71.7/89.3/84.5 & 97.2/99.0/94.7 & \textbf{99.7/99.9/99.2} & 98.3/99.3/98.4  & 94.6/97.9/95.9 \\
Transistor & 78.2/79.5/68.8  & 94.2/95.2/90.0 & 99.8/99.6/97.4 & \textbf{100.0/100.0/100.0} & 98.3/99.3/96.8 \\
Zipper & 88.4/96.3/93.1 & 99.5/99.9/99.2 & 95.1/99.1/94.4 & 99.3/99.8/97.5  & \textbf{100.0/99.9/99.9} \\
Carpet & 95.9/98.8/94.9 & 98.5/99.6/97.2 & 99.4/99.9/98.3 & \textbf{99.8/99.9}/\textbf{99.4}  & 99.1/99.8/97.9 \\
Grid & 97.9/99.2/96.6 & 98.0/99.4/96.5 & 98.5/99.8/97.7 & \textbf{100.0/100.0/100.0}  & \textbf{100.0/100.0/100.0} \\
Leather & 99.2/99.8/98.9 & \textbf{100.0/100.0/100.0}& 99.8/99.7/97.6 & \textbf{100.0/100.0/100.0}  & \textbf{100.0/100.0/100.0} \\
Tile & 97.0/98.9/95.3 & 98.3/99.3/96.4 & 96.8/99.9/98.4 & 98.2/99.3/95.4  & \textbf{98.6/99.5/96.5} \\
Wood & 99.9/100.0/99.2 & 99.2/99.8/98.3 & \textbf{99.7/100.0/100.0} & 98.8/99.6/96.6 & 99.4/99.8/98.3 \\
\midrule
Mean & 89.2/95.5/91.6 & 94.6/96.5/95.2 & 97.2/99.0/96.5 & \textbf{98.6/99.6}/97.8 & \textbf{98.6}/99.5/\textbf{97.9} \\
\bottomrule
\end{tabular}
\label{tab:det}
\end{center}
   \vspace{-1em}

\end{table*}

\section{Experiments}
\subsection{Setups: Dataset, Metrics, and Implementation}
\textbf{Dataset.}
The \textbf{MVTec-AD}~\cite{bergmann2019mvtec} dataset contains 5354 images divided into 3629 training defects-free images and 1725 test images (w and w/o defects).
Images belonging to 15 different classes come with pixel-wise anomaly annotations. 


\textbf{Metrics.} We report on the Area Under the Receiver Operating
Characteristic Curve (AU-ROC), Average Precision (AP), F1-score-max (\(F1_{\text{max}}\)), and Area Under the Per-Region-Overlap  (AU-PRO) for anomaly detection and localization.

\begin{table*}[htbp]
\caption{Comparison with SoTA methods on \textbf{MVTec-AD} dataset for multi-class anomaly localization with AU-ROC/AP/$F1_{\text{max}}$/AU-PRO.}
\begin{center}
\begin{tabular}{c|ccccc}
\toprule
\textbf{Object} & \textbf{DeSTeg~\cite{Zhang2023}} & \textbf{RD4AD~\cite{Deng2022}} & \textbf{DiAD~\cite{He2024}} & \textbf{MambaAD~\cite{He2024arXiv}} & \textbf{Pyramid-Mamba~(Ours)} \\
\midrule
Bottle & 97.2/53.8/62.4/89.0 & 97.8/68.2/67.6/94.0 & 98.4/52.2/54.8/86.6 & 98.8/79.7/76.7/95.2 & \textbf{98.9/81.0/77.4/96.1} \\
Cable & 96.7/42.4/51.2/85.4 & 85.1/26.3/33.6/75.1 & 96.8/50.1/57.8/80.5 & 95.8/42.2/48.1/\textbf{90.3} & \textbf{96.8/42.9/48.9}/90.1 \\
Capsule & 98.5/5.4/44.3/84.5 & 98.8/43.4/50.0/94.8 & 97.1/42.0/45.3/87.2 & 98.4/43.9/47.7/92.6 & \textbf{98.6/45.2/48.3/94.0} \\
Hazelnut & 98.4/44.6/51.4/87.4 & 97.9/36.2/51.6/92.7 & 98.3/79.2/80.4/91.5 & 99.0/63.6/64.4/95.7 & \textbf{99.2/70.2/68.8/95.6} \\
Metal Nut & 98.0/83.1/79.4/85.2 & 94.8/55.5/66.4/91.9 & 97.3/30.0/38.3/90.6 & 96.7/74.5/79.1/93.7 & \textbf{97.1/77.3/80.2/93.9} \\
Pill & 96.5/72.4/67.7/81.9 & 97.5/63.4/65.2/95.8 & 95.7/46.0/51.4/89.0 & 97.4/64.0/66.5/95.7 & \textbf{97.6/65.8/67.4/97.1} \\
Screw & 96.5/15.9/23.2/84.0 & 99.4/40.2/44.6/96.8 & 97.9/60.6/59.6/95.0 & 99.5/49.8/50.9/97.1 & \textbf{99.5/51.1/51.0/97.6} \\
Toothbrush & 98.4/46.9/52.5/87.4 & 99.0/53.6/58.8/92.0 & \textbf{99.0/78.7/72.8/95.0} & 99.0/48.5/59.2/91.7 & 99.0/49.1/60.8/92.5 \\
Transistor & 95.8/58.2/56.0/83.2 & 85.9/42.3/45.2/74.7 & 95.1/15.6/31.7/90.0 & 96.5/69.4/\textbf{67.1}/87.0 & \textbf{96.9/69.7}/67.0/\textbf{91.6} \\
Zipper & 97.9/53.4/54.6/90.7 & \textbf{98.5}/53.9/60.3/94.1 & 96.2/60.7/60.0/91.6 & 98.4/\textbf{60.4/61.7}/94.3 & 98.3/59.1/61.2/\textbf{94.6} \\
Carpet & 97.4/38.7/43.2/90.6 & 99.0/58.5/60.4/95.1 & 98.6/42.2/46.4/90.6 & 99.2/60.0/63.3/96.7 & \textbf{99.3/65.1/64.5/97.2} \\
Grid & 96.8/20.5/27.6/88.6 & 96.5/23.0/28.4/97.0 & 97.0/42.1/46.9/86.8 & 99.2/47.4/47.7/97.0 & \textbf{99.2/49.1/48.5/97.5} \\
Leather & 98.7/28.5/32.9/92.7 & 99.3/38.0/45.1/97.4 & 98.8/56.1/62.3/91.3 & 99.4/50.3/53.3/98.7 & \textbf{99.4/50.4/53.5/98.7} \\
Tile & 95.7/60.5/59.9/90.6 & 95.3/48.5/60.5/85.8 & 92.4/65.7/64.1/90.7 & 93.8/45.1/54.8/80.0 & \textbf{93.8/45.5/55.0/81.7} \\
Wood & 91.4/34.8/39.7/76.3 & 95.3/47.8/51.0/90.0 & 93.3/43.3/43.5/97.5 & 94.4/46.2/48.2/91.2 & \textbf{94.7/48.3/49.4/99.8} \\
\midrule
Mean & 96.9/45.9/49.7/86.5 & 96.1/48.6/53.8/91.1 & 96.8/52.6/55.5/90.7 & 97.7/56.3/59.2/93.1 & \textbf{97.8/57.3/59.7/93.5} \\
\bottomrule
\end{tabular}
\label{tab:loc}
\end{center}
   \vspace{-1em}

\end{table*}

\textbf{Implementation Details.}
We conducted the experiments considering a $256 \times 256$ input images with no further augmentation
We adopted a ResNet34 feature encoder.
Our decoder has $[3,4,6,3]$ CSS modules for the corresponding encoder layers and works with $M=3$.
We used $P=2$ pyramid layers.
To mitigate the lack of anomalous images during training, we add random noise to the normal images only.
Our loss function is the sum of MSE at the multiple scales.
We optimize our model using the Adam optimizer with a decay rate of 1e-4 and a learning rate of 5e-4 for 500 epochs.

\subsection{Comparison with State-of-the-art Methods}

Table \ref{tab:det} and \ref{tab:loc} present a comprehensive comparison of our proposed model against state-of-the-art methods for multi-class anomaly detection and localization on the MVTec-AD dataset.
Our Pyramid-Mamba demonstrates competitive performance across various object classes, achieving the highest mean scores in AU-ROC ($98.6\%$, tied with MambaAD) and $F1_{\text{max}}$ ($97.9\%$).
Our method outperforms or matches the best results in several categories also showing perfect detection ($100\%$ across all metrics) for classes such as Bottle, Hazelnut, Grid, and Leather, thus highlighting its robustness across different types of anomalies.
We achieve superior localization performance across all mean metrics than existing models thus setting a new state-of-the-art result.
We demonstrate consistent improvements over the previous best method, MambaAD \cite{Huang2024}, particularly in AP ($+1.0\%$), $F1_{\text{max}}$ ($+0.5\%$), and AU-PRO ($+0.4\%$).
Notably, our method outperforms or matches the best results in 12 out of 15 object categories for AU-ROC, showcasing its robustness across diverse anomaly types.
Significant improvements are observed in challenging categories such as Hazelnut (AP: $+6.6\%$, $F1_{\text{max}}$: $+4.4\%$) and Metal Nut (AP: $+2.8\%$, $F1_{\text{max}}$: $+1.1\%$).
The enhanced localization performance can be attributed to our model's hierarchical structure, which enables efficient capture of both local and global contexts. By processing image features at multiple scales, our approach effectively identifies fine-grained anomalies while maintaining global coherence, addressing a key limitation of previous methods. 

\begin{figure}[t]
  \centering
   \includegraphics[width=1\linewidth]{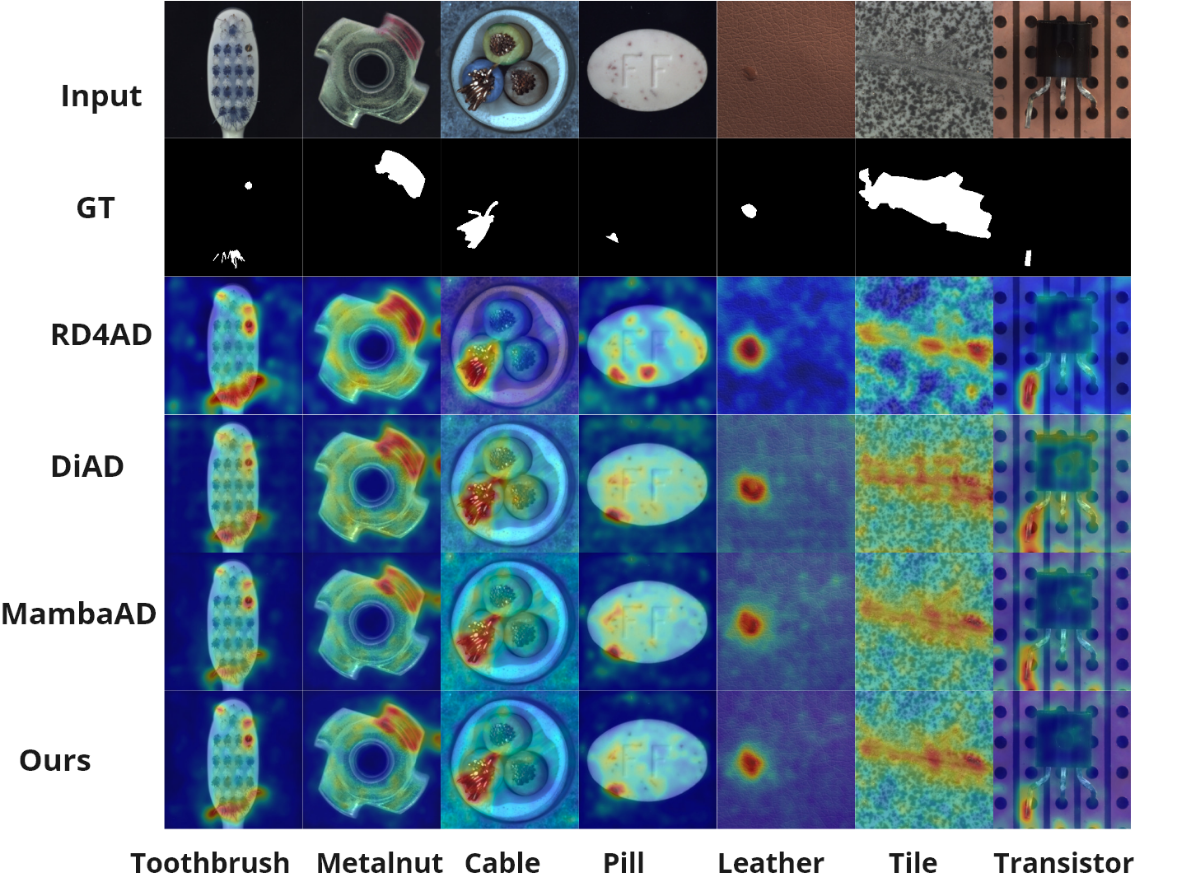}
   \caption{Qualitative visualization for pixel-level anomaly segmentation on MVTec dataset.}
   \label{fig:example}
   \vspace{-1.5em}
\end{figure}

\textbf{Qualitative Results.}
Extensive experiments conducted on MVTec AD dataset demonstrate the effectiveness of our approach in achieving SoTA performance.  The qualitative evaluation of performance, as visualized in ~\figurename~\ref{fig:example} demonstrates our model’s versatility and
effectiveness in detecting and localizing anomalies across
a wide range of objects.

\subsection{Ablation Study}
Through the ablation study, we want to answer different questions that would help us understand the importance of each proposed component of our architecture.
\begin{table}[htbp]
  \caption{Comparison of mean performance values for anomaly detection and localization tasks across different ablations.}
  \begin{center}
  
  \begin{tabularx}{\linewidth}{l|C|C}
    \toprule
    \textbf{Category} & \textbf{MVTec Anomaly
Detection} (AU-ROC/AP/$F1_{\text{max}}$) & \textbf{MVTec Anomaly
Localization} (AU-ROC/AP/$F1_{\text{max}}$/AU-PRO) \\
    \midrule
    w/o $\mathbf{X}_{\text{g}}$        & 98.4/99.4/97.5  & 97.7/56.2/59.0/93.4  \\
    w/o $\mathbf{X}_{\text{c},k}$      & 98.5/99.4/97.7  & 97.8/56.7/59.1/93.4  \\
    Pyramid-Mamba              & \textbf{98.6}/\textbf{99.5}/\textbf{97.9}  & \textbf{97.8}/\textbf{57.3}/\textbf{59.7}/\textbf{93.5}  \\
    \bottomrule
  \end{tabularx}
  \label{tab:local_vs_global_analysis}
  \end{center}
\end{table}

\begin{table*}[htbp]
\caption{Mean performance values for anomaly detection and localization tasks on the \textbf{MVTec-ad} dataset with various types of artificial noise.}
\begin{center}
\scriptsize
\begin{tabularx}{\linewidth}{c|CCCC}
\toprule
\textbf{Task} & \textbf{No Pyramid, No Noise} & \textbf{No Pyramid, Noise} & \textbf{Pyramid, No Noise} & \textbf{Pyramid, Noise} \\
\midrule
Detection (AU-ROC/AP/$F1_{\text{max}}$) & 98.6/99.6/97.8 & 98.5/99.5/97.8 & 98.5/99.6/97.9 & 98.6/99.5/97.9 \\
Localization (AU-ROC/AP/$F1_{\text{max}}$/AU-PRO) & 97.7/56.3/59.2/93.1 & 97.8/57.1/59.6/93.4 & 97.8/56.9/59.1/93.5 & 97.8/57.3/59.7/93.5 \\
\bottomrule
\end{tabularx}
\label{tab:noise_combined}
\end{center}
   \vspace{-1em}

\end{table*}


\textbf{What is the importance of local and global features?}
Table~\ref{tab:local_vs_global_analysis} highlights the significance of pyramid global features ($\mathbf{X}_{\text{g}}$) in both anomaly detection and localization tasks.
Removing $\mathbf{X}_{\text{g}}$ leads to a 
$0.4\%$ $F1_{\text{max}}$ drop in detection and a 
$1.9\%$ reduction in AP for localization, demonstrating that pyramid global features are critical for capturing multi-scale information and enhancing anomaly localization. This performance gap underscores how the pyramid scanning mechanism effectively models anomalies across different scales, significantly improving both precision and robustness. 
Similarly, while less impactful, removing local features ($\mathbf{X}_{\text{c}, k}$) results in performance degradation, validating the necessity of fine-grained local context.
The results confirm that both global features and local features are essential for accurate, multi-scale anomaly detection and localization.

\begin{figure}[t]
  \centering
   \includegraphics[height=0.2\textheight]{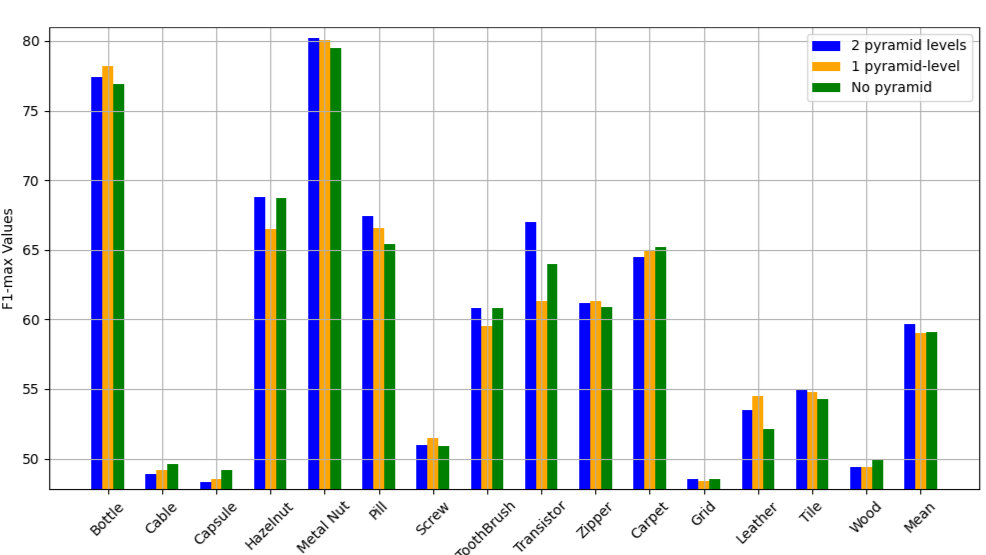}
   \caption{Ablation studies on multiclass anomaly localization for different pyramid levels. Results are on the MVTec dataset.}
   \label{fig:pyramid_levels_analysis}
      \vspace{-1em}

\end{figure}
\textbf{What is the impact of the pyramid levels?}
Results depicted in ~\figurename~\ref{fig:pyramid_levels_analysis} show the performance of our method across multiple object categories for multiclass anomaly localization, using the $F1_{\text{max}}$ metric under different pyramid levels.
Utilizing 2 pyramid levels (blue bars) consistently enhances performance, yielding the highest $F1_{\text{max}}$ scores across most categories compared to the single pyramid level (orange bars) and the no pyramid (green bars) configurations.
Categories like \textit{Hazelnut}, \textit{Metal Nut}, and \textit{Zipper} exhibit a substantial boost in $F1_{\text{max}}$ when pyramid-based processing is applied.

This performance gap reinforces the importance of multi-scale modeling introduced by the pyramid structure, as it captures anomalies at various granularities. The mean performance shows a clear trend of improved $F1_{\text{max}}$ values with 2 pyramid levels, further validating the necessity of hierarchical processing for robust anomaly localization.

\textbf{Effect of adding synthetic anomalies}
Table~\ref{tab:noise_combined} presents the mean performance metrics for anomaly detection and localization tasks, evaluated across various noise conditions and methods.
For anomaly detection, the highest mean AU-ROC/AP/$F1_{\text{max}}$ values are observed when the pyramid approach is used together with the feature level noise, with scores approaching 98.6/99.5/97.9 for the three considered metrics.
Similar results are obtained for anomaly localization.
Notably, the AP performance drops by about $1\%$ when the pyramid and the noise are not jointly considered, indicating potential vulnerabilities in localization tasks when not considering perturbations and multiple feature scales.

\section{Conclusion}
We introduced a novel Pyramid-based Mamba for multi-class anomaly detection and localization. The proposed method includes a pre-trained encoder, a synthetic anomaly generator and a novel pyramid-based Mamba decoder. Within each Mamba decoder, the CSS module contains a PSS that performs a novel pyramidal scanning, integrating the long-range modeling capabilities of Mamba at different pyramid levels.
Additionally, the CSS module includes an LEC block that extracts local features using parallel multi-kernel convolution operation. The combination of the PSS and LEC enables the model to capture both local and global features, contributing to the detection of fine-grained anomalies. Comprehensive experiments on the MVTec-AD benchmark demonstrated that our method achieves superior performance than existing methods at both image and pixel levels.


\bibliographystyle{IEEEbib}
\bibliography{icme2025references}

\begin{thebibliography}{10}

\bibitem{Defard2021}
Thomas Defard, Aleksandr Setkov, Angelique Loesch, and Romaric Audigier,
\newblock ``Padim: A patch distribution modeling framework for anomaly detection and localization,''
\newblock in {\em ICPR}, 2021, pp. 475--489.

\bibitem{Roth2022}
Karsten Roth, Latha Pemula, Joaquin Zepeda, Bernhard Sch{\"o}lkopf, Thomas Brox, and Peter Gehler,
\newblock ``Towards total recall in industrial anomaly detection,''
\newblock in {\em CVPR}, 2022, pp. 14318--14328.

\bibitem{Li2021}
Chun-Liang Li, Kihyuk Sohn, Jinsung Yoon, and Tomas Pfister,
\newblock ``Cutpaste: Self-supervised learning for anomaly detection and localization,''
\newblock in {\em CVPR}, 2021, pp. 9664--9674.

\bibitem{He2024}
H.~He, J.~Zhang, H.~Chen, X.~Chen, Z.~Li, X.~Chen, Y.~Wang, C.~Wang, and L.~Xie,
\newblock ``A diffusion-based framework for multi-class anomaly detection,''
\newblock in {\em AAAI}, 2024.

\bibitem{Deng2022}
H.~Deng and X.~Li,
\newblock ``Anomaly detection via reverse distillation from one-class embedding,''
\newblock in {\em CVPR}, 2022.

\bibitem{You2022}
Z.~You, L.~Cui, Y.~Shen, K.~Yang, X.~Lu, Y.~Zheng, and X.~Le,
\newblock ``A unified model for multi-class anomaly detection,''
\newblock in {\em NeurIPS}, 2022.

\bibitem{Liu2024}
Yue Liu, Yunjie Tian, Yuzhong Zhao, Hongtian Yu, Lingxi Xie, Yaowei Wang, Qixiang Ye, and Yunfan Liu,
\newblock ``Vmamba: Visual state space model,''
\newblock {\em arXiv preprint arXiv:2401.10166}, 2024.

\bibitem{Shi2024}
Y.~Shi, B.~Xia, X.~Jin, X.~Wang, T.~Zhao, X.~Xia, X.~Xiao, and W.~Yang,
\newblock ``Vmambair: Visual state space model for image restoration,''
\newblock {\em arXiv preprint arXiv:2403.11423}, 2024.

\bibitem{Ruan2024}
J.~Ruan and S.~Xiang,
\newblock ``Vm-unet: Vision mamba unet for medical image segmentation,''
\newblock {\em arXiv preprint arXiv:2402.02491}, 2024.

\bibitem{Wang2024}
Z.~Wang, J.-Q. Zheng, Y.~Zhang, G.~Cui, and L.~Li,
\newblock ``Mamba-unet: Unet-like pure visual mamba for medical image segmentation,''
\newblock {\em arXiv preprint arXiv:2402.05079}, 2024.

\bibitem{Huang2024}
T.~Huang, X.~Pei, S.~You, F.~Wang, C.~Qian, and C.~Xu,
\newblock ``Localmamba: Visual state space model with windowed selective scan,''
\newblock {\em arXiv preprint arXiv:2403.09338}, 2024.

\bibitem{Gu2023}
A.~Gu and T.~Dao,
\newblock ``Mamba: Linear-time sequence modeling with selective state spaces,''
\newblock {\em arXiv preprint arXiv:2312.00752}, 2023.

\bibitem{Gong2019}
Dong Gong, Lingqiao Liu, Vuong Le, Budhaditya Saha, Moussa~Reda Mansour, Svetha Venkatesh, and Anton van~den Hengel,
\newblock ``Memorizing normality to detect anomaly: Memory-augmented deep autoencoder for unsupervised anomaly detection,''
\newblock in {\em ICCV}, 2019, pp. 1705--1714.

\bibitem{Zavrtanik2021}
Vitjan Zavrtanik, Matej Kristan, and Danijel Sko{\v c}aj,
\newblock ``Reconstruction by inpainting for visual anomaly detection,''
\newblock {\em Pattern Recognition}, vol. 112, pp. 107706, 2021.

\bibitem{Lu2023}
R.~Lu, Y.~Wu, L.~Tian, D.~Wang, B.~Chen, X.~Liu, and R.~Hu,
\newblock ``Hierarchical vector quantized transformer for multi-class unsupervised anomaly detection,''
\newblock in {\em NeurIPS}, 2023.

\bibitem{Gu2021}
Albert Gu, Karan Goel, and Christopher R{\'e},
\newblock ``Efficiently modeling long sequences with structured state spaces,''
\newblock in {\em ICLR}, 2021.

\bibitem{Smith2022}
J.~T. Smith, A.~Warrington, and S.~W. Linderman,
\newblock ``Simplified state space layers for sequence modeling,''
\newblock in {\em ICLR}, 2022.

\bibitem{Mehta2022}
H.~Mehta, A.~Gupta, A.~Cutkosky, and B.~Neyshabur,
\newblock ``Long range language modeling via gated state spaces,''
\newblock in {\em ICLR}, 2022.

\bibitem{Zhang2023}
X.~Zhang, S.~Li, X.~Li, P.~Huang, J.~Shan, and T.~Chen,
\newblock ``Destseg: Segmentation guided denoising student-teacher for anomaly detection,''
\newblock in {\em CVPR}, 2023.

\bibitem{He2024arXiv}
H.~He, Y.~Bai, J.~Zhang, Q.~He, H.~Chen, Z.~Gan, L.~Wang, Y.~Yang, X.~Li, Z.~Dou, and L.~Xie,
\newblock ``Mambaad: Exploring state space models for multi-class unsupervised anomaly detection,''
\newblock in {\em NeurIPS}, 2024.

\bibitem{bergmann2019mvtec}
Paul Bergmann, Michael Fauser, David Sattlegger, and Carsten Steger,
\newblock ``Mvtec ad--a comprehensive real-world dataset for unsupervised anomaly detection,''
\newblock in {\em CVPR}, 2019, pp. 9592--9600.

\end{thebibliography}

\end{document}